\documentclass{article}

\usepackage{times}
\usepackage{graphicx}
\usepackage{subfigure}
\usepackage[square,sort,comma,numbers]{natbib}

\usepackage{algorithm}
\usepackage{algorithmic}
\usepackage{amsmath}
\usepackage{graphicx}
\usepackage{color}
\graphicspath{ {graphs/} }

\DeclareMathOperator*{\argmin}{argmin}

\usepackage[preprint,nonatbib]{nips_2018}

\usepackage[utf8]{inputenc} 
\usepackage[T1]{fontenc}    
\usepackage{hyperref}       
\usepackage{url}            
\usepackage{booktabs}       
\usepackage{amsfonts}       
\usepackage{nicefrac}       
\usepackage{microtype}      
\usepackage{caption}

\title{Tight Prediction Intervals Using Expanded Interval Minimization}

\author{
  Dongqi Su \\
  GoDaddy, Inc. \\
  \texttt{dsu5@godaddy.com} \\
  \And
  Ying Yin Ting \\
  GoDaddy, Inc. \\
  \texttt{yying@godaddy.com} \\
  \And
  Jason Ansel \\
  GoDaddy, Inc. \\
  \texttt{jansel@godaddy.com} \\
}

\begin{document}

\maketitle

\vspace{-1em}

\begin{abstract}

Prediction intervals are a valuable way of quantifying uncertainty in regression problems.  Good prediction intervals should be both correct, containing the actual value between the lower and upper bound at least a target percentage of the time; and tight, having a small mean width of the bounds. Many prior techniques for generating prediction intervals make assumptions on the distribution of error, which causes them to work poorly for problems with asymmetric distributions.

This paper presents Expanded Interval Minimization (EIM), a novel loss function for generating prediction intervals using neural networks.  This loss function uses minibatch statistics to estimate the coverage and optimize the width of the prediction intervals.  It does not make the same assumptions on the distributions of data and error as prior work. We compare to three published techniques and show EIM produces on average $1.37x$ tighter prediction intervals and in the worst case $1.06x$ tighter intervals across two large real-world datasets and varying coverage levels.

\end{abstract} 

\section{Introduction} \label{introduction}

Prediction intervals are the preferred method of many people for quantifying uncertainty~\cite{halberg2009maximum}. They work by providing a lower and upper bound for an estimated variable such that the value of the predicted variable falls between the upper and lower bound for at least some target percentage (e.g., 90\%) of holdout data.  One can trivially achieve this correctness criterion with the bounds $[-\infty, \infty]$, so to be useful one also wants prediction intervals that are {\em tight}, having the minimum possible mean bound width while still satisfying the correctness criteria.  There are three published techniques for generating prediction intervals with neural networks which we use as baselines to evaluate EIM.

Maximum likelihood estimation~\cite{374138,nix1995learning} is a well-known method that can build prediction intervals based on two neural networks, one predicting the value and the second predicting the error.   Ensemble method~\cite{heskes1997practical,carney1999confidence} is another way of generating prediction intervals using variances among ensembles of many models to estimate error. Both of these techniques work well for problems with symmetric distributions because a Gaussian assumption on model error motivates their designs; however, they struggle to produce optimal bounds for the asymmetric distributions found in our datasets.

Quantile regression method~\cite{koenker2001quantile, koenker1978regression} is a modified version of least squares that converges to a given quantile of a dataset.  One can use quantile regression to build prediction intervals that span between two given quantiles.  For example, one could generate a prediction interval that spans from the 10\%  quantile regression as a lower bound to the 90\% quantile regression as an upper bound.  We set these two upper and lower quantiles via exhaustive grid search to find the optimal values for each target coverage.  This technique is more flexible than the prior two, in that it does not assume a specific data distribution; however, it is limited in that it is only able to express intervals in a quantile-to-quantile form.

This paper presents a novel technique for generating prediction intervals with neural networks called {\em Expanded Interval Minimization} (EIM).  We build a neural network structure that outputs both a lower and upper bound directly. The network uses a loss function that first scales the output bounds to cover the given target percentage of the minibatch, and then minimizes the width of those scaled bounds during training.  The scaling ensures that {\em correctness criteria} is met on the training data and then the training process optimizes only the {\em tightness} of those already correct bounds.  Unlike maximum likelihood method and ensemble method, EIM can support asymmetric data and error distributions, and it is also able to create prediction intervals varying in the predicted quantiles between different samples.  We empirically demonstrate that EIM produces tighter prediction intervals than prior techniques, both on the real-word Domain Valuation Dataset and the Million Song Dataset. Compared to the next best method, EIM produces at least $1.26x$ tighter prediction intervals for a 70\% coverage target, $1.21x$ tighter prediction intervals for an 80\% target, and $1.06x$ tighter intervals for a 90\% coverage target.








\section{Related Work}

Some past surveys focus on comparing techniques to construct the prediction interval for the neural network point estimation~\cite{khosravi2011comprehensive,zapranis2005prediction}.
The most common techniques to construct the prediction interval are the delta method (also known as analytical method)~\cite{khosravi2011comprehensive,zapranis2005prediction,de1998prediction,hwang1997prediction},
methods that directly predict the variance (maximum likelihood method and ensemble method) ~\cite{374138,zapranis2005prediction,nix1995learning, khosravi2011comprehensive,heskes1997practical,carney1999confidence} and quantile regression method ~\cite{koenker1978regression, koenker2001quantile}. 
Our baseline methods exclude the delta method because it is a poor fit for our datasets. The delta method assumes constant error variance in the dataset and noise homogeneity in all the samples, while our datasets have substantial differences in variance between different types of samples. We implement maximum likelihood method, ensemble method, quantile regression method and a naive baseline fixed bounds method to compare with our proposed technique EIM.


The prior work maximum likelihood method~\cite{374138,nix1995learning} and ensemble method~\cite{heskes1997practical,carney1999confidence}, make an assumption that the target distribution ($y(x)$) can be broken into two independent terms representing the true regression and the noise.  This assumption leads to these techniques estimating the {\em total prediction variance}:
    \begin{equation}
        \sigma_p^2(x) = \sigma_m^2(x) + \sigma_\epsilon^2(x)
    \end{equation}
where $\sigma_m^2(x)$ is the model uncertainty variance and $\sigma_\epsilon^2(x)$ is the data noise variance. These methods then add and subtract a constant $k$ times the estimate of $\sigma_p(x)$ from the estimated regression $f(x)$ to construct the prediction intervals:
\begin{equation}
[f(x) - k{\sigma}_p(x), f(x) + k{\sigma}_p(x)]
\end{equation}

\subsection{Maximum Likelihood Method} \label{maximum_likelihood_method}
Maximum likelihood method~\cite{374138,nix1995learning} builds two neural networks.  The first neural network estimates the regression and the second neural network estimates the {total prediction variance} $\sigma_p^2(x)$. The loss function $E_y$ for the first neural network, is simply mean squared error:
    \begin{equation} \label{mse}
        E_t = \sum_{i=1}^N (y_i - f_{mle}(x_i))^2
    \end{equation}
where $y_i$ is the target value correspond to $x_i$ and $f_{mle}(x_i)$ is the output of the first neural network. The loss function $E_{\sigma_p^2}$, for the second neural network is the mean squared error of the squared error of the first neural network:
    \begin{equation}
        E_{\sigma_p^2} = \sum_{i=1}^N \left(
        \left( y_i - f_{mle}(x_i) \right)^2 - var_{mle}(x_i)
        \right)^2
    \end{equation}
where $var_{mle}(x_i)$ is the output of the second neural network. The final prediction interval is [$f_{mle}(x_i) - k \sqrt{var_{mle}(x_i)}$, $f_{mle}(x_i) + k \sqrt{var_{mle}(x_i)}$], where $k$ depends on the desired level of confidence motivated by the Gaussian distribution. Following the recommendations of the authors, we train these two neural networks by splitting our training data in half using disjoint training data for each network.

\subsection{Ensemble Method} \label{ensemble_method}

Ensemble method~\cite{heskes1997practical,carney1999confidence} tries to estimate $\sigma_p^2(x)$ by estimating $\sigma_m^2(x)$ and $\sigma_\epsilon^2(x)$ separately and then combining them. It prepares M bootstrap neural networks each trained on a subsample of the full dataset using \eqref{mse}. It estimates the regression by taking the average output of all $M$ bootstrap neural networks. We denoted this estimated value as $f_{emb}(x)$

To estimate the model uncertainly $\sigma_m^2(x)$, it splits all $M$ bootstrap neural network into $M_2$ group and calculated an initial set $L$ of $M_2$ average predictions for each group. Given this initial set $L$, it then estimates $\sigma_m^2(x)$ by taking the average of variances of $P$ bootstrap set sampled with replacement from the initial set $L$. We denoted this estimated value as $\hat{\sigma}_{emb}^2(x)$.

Given $f_{emb}(x)$ and $\hat{\sigma}_{emb}^2(x)$, a new model is trained on the same dataset to estimate $\sigma_\epsilon^2(x)$ directly in quantity $\hat{\sigma}_{\epsilon}^2(x)$ using maximum likelihood ~\cite{374138} with the assumption that $\epsilon(x) \sim \mathcal{N}(0, \sigma_{\epsilon}^2(x))$. For observed residual $r$, the loss function is therefore the negative probability density function of $\epsilon(x)$, as follows:
    \begin{equation} \label{normal_pdf_loss}
        E_{\sigma_{\epsilon}^2} = - \sum_{i=1}^{N} log(\frac{1}{{\sqrt {2\pi \hat{\sigma}_\epsilon^2(x_i)}}} exp(-\frac{r_i^2}{2\hat{\sigma}_{\epsilon}^2(x_i)}))
    \end{equation}
where $r_i^2 = (y_i-f_{emb}(x_i))^2 - \hat{\sigma}_{emb}^2(x_i)$, in which we need to remove the uncertainty generated by the estimated true regression. Similar to maximum likelihood method, ensemble method constructs prediction intervals using the estimated total prediction variance by summing $\hat{\sigma}_{emb}^2(x)$ and $\hat{\sigma}_{\epsilon}^2(x)$ and the estimated true regression $f_{emb}(x)$. For our implementation, we used $M=200, M_2=8, P=1000$ as in the original paper. We applied the log operation to cancel out the exponential term in \eqref{normal_pdf_loss} before we used it as a loss function to prevent exploding gradients.

\subsection{Quantile Regression Method} \label{quantile_regression}

Quantile regression~\cite{koenker2001quantile, koenker1978regression} is similar to the ordinary least squares regression, however unlike the ordinary least square regression that estimates the mean of conditional target distribution, quantile regression estimates the $\tau$ quantile of the conditional target distribution. For example, the 50th quantile of a target distribution is the conditional median of that target distribution, which is above 50\% of the target values. To train a neural network that estimates the $\tau$ quantile of the target distribution, quantile regression uses the loss function $E_{\tau}$ according to:
    \begin{equation}
        E_{\tau} = \sum_{i=1}^{N} L_{\tau}(y(x_i) - q_{\tau}(x_i))
        \quad
        \text{where}
        \quad
        L_{\tau}(e_i) = 
        \begin{cases}
            \tau e_i & \text{if $e_i \geq 0$} \\
            (\tau - 1) e_i & \text{otherwise} \\
        \end{cases}
    \end{equation}
where the output of the model is $q_{\tau}(x)$. Given a desired quantile range from $\tau_l$ to $\tau_u$, we construct a prediction interval by training two quantile regression models: one for the lower bound $\tau_l$ and one for the upper bound $\tau_u$. For our implementation, we use a single neural network that output both the lower and upper quantile regressions. 

A naive heuristic to set the hyperparameters $\tau_l$ and $\tau_u$ would be, for example, to use 10\% and 90\% to capture 80\% of target coverage. Unfortunately, our experiment shows that this produces sub-optimal results. Instead, we set $\tau_l$ and $\tau_u$ via exhaustive grid search at training time.

\subsection{Fixed Bounds Method} \label{sec:fixed}

As a naive baseline, we include a fixed bounds method in our experiments. One can view the output of this model as a ceiling, or the score one could trivially get without using any advanced technique. This technique takes a model that is trained to predict a regression ($f(x)$) and creates prediction intervals that are plus or minus a fixed percentage ($\alpha$) of the target value, yielding: 
    \begin{equation}
    [(1 - \alpha) f(x), (1 + \alpha) f(x)]
    \end{equation}

\section{Assessment Metrics} \label{assessment_metrics}

Assessing the quality of a prediction interval can be difficult because there are two competing objectives.  One wants intervals that are both {\em correct} the target amount of time and {\em tight}, having a narrow mean width.  These two objectives were formalized in~\cite{shrestha2006machine} as prediction interval coverage probability (PICP) and mean prediction interval width (MPI~\cite{shrestha2006machine} or MPIW~\cite{khosravi2011comprehensive, khosravi2010construction, khosravi2010prediction}).

With prediction interval coverage probability (PICP) representing the percentage of the time the prediction interval is correct:
    \begin{equation}
        \text{PICP}_{l(x), u(x)} = \frac{1}{N}\sum_{i=1}^{N} h_i 
        \quad
        \text{where}
        \quad
        h_i = \begin{cases}
        1 & \text{if } l(x_i) \leq y_i \leq u(x_i) \\
        0 & \text{otherwise}
    \end{cases}
    \end{equation}
where $N$ is the size of the test set. $l(x_i)$ and $u(x_i)$ are the lower and upper bounds of the prediction interval for sample $i$, and $y_i$ is the observed target value. Mean prediction interval width (MPIW), measures the average size of all prediction intervals. 
    \begin{equation}
    \text{MPIW}_{l(x), u(x)} = \frac{1}{N}\sum_{i=1}^{N} | u(x_i) - l(x_i) |
    \end{equation}
Given Target $T$, one wants to minimize $\text{MPIW}$, while maintaining $\text{PICP} \geq T$. Some literature discards cases where $\text{PICP} < T$ as invalid because they fail to meet the correctness criteria of the problem.  \cite{khosravi2011lower}~propose a combined metric that applies an exponentially exploding penalty to cases that do not meet the PICP target.  In practice, if the penalty is sufficiently large, this eliminates incorrect intervals, but one can construct still pathological datasets to defeat this combined metric. Another way to address this issue, which we use in evaluation (see section \ref{scaledintervals}), linearly scales the output of each technique's output by a constant factor to make it hit the target PICP exactly. 

\section{Expanded Interval Minimization} \label{EIM}

To motivate the design of Expanded Interval Minimization (EIM), imagine how one would find the parameters ($\theta$) of a neural network to output a prediction interval lower bound $l(x; \theta)$, and an upper bound $u(x; \theta)$ that optimizes the objective directly:
    \begin{equation}
        \theta_{min} = \argmin_{\theta} \text{MPIW}_{l(x; \theta), u(x; \theta)}
        \quad
        \text{subject to}
        \quad
        \text{PICP}_{l(x; \theta), u(x; \theta)} = T
    \end{equation}
This formulation would be desirable, but it is difficult in practice because it is expensive and difficult to optimize directly.

A critical insight is that one can use the PICP and MPIW of each minibatch as a noisy estimate of the population PICP and MPIW. Calculating MPIW in a minibatch is straightforward, but the minibatch PICP is unlikely to match the target $T$ during training. We fix this issue by applying automatic scaling of the bounds so they cover the target PICP on the minibatch.  For each minibatch ($B$), we calculate a scaling factor $k_B$ to expand or shrink the predicted bounds such that the minibatch PICP equals $T$ throughout the training. The EIM loss function ($E$) for minibatch $B$, is as follows:
    \begin{equation}
        E_B = k_B \sum_{i \in B} \left| u(x_i) - l(x_i) \right|
    \end{equation}
    
Here we present a closed form solution to calculate $k_B'$, the scaling required to hit the PICP target on the minibatch.  $k_B'$ is a simplified version of the full $k_B$ which will be introduce later.
First, we to calculate a set of minimum scaling factors $\{k_i\}_{i=1}^{|B|}$ for all instances in the minibatch such that each scaled output bound is just wide enough to capture the target value $y_i$, according to:
    \begin{equation}
        k_{i} = 
        \left| 
        \frac{
        u(x_i) + l(x_i) - 2y_i
        }{
        u(x_i) - l(x_i)
        } \right|
    \end{equation}

Then we select the $T$th percentile\footnote{``$T$th percentile'' represents selecting the value $T$ percent of the way through the set when sorted by value.  In this context we interpret $T$ as a percentage between 0 and 100, while in other places it is 0 to 1.} scaling factor from  $\{k_i\}_{i=1}^{|B|}$ to scale all the output bounds in minibatch $B$. This will make the output bounds in the minibatch achieve exactly $T$ PICP.
    \begin{equation}
        k_{B}' = \sum_{i \in B} c_i k_i
    \quad
    \text{where} 
    \quad
    c_i = 
    \begin{cases}
        1 & \text{if $k_i$ is the $T$th percentile of $B$} \\
        0 & \text{otherwise} \\
    \end{cases}
    \end{equation}

While the above loss function works and directly represents our primary objective, it produces suboptimal results since only a single value from the minibatch receive nonzero gradients from the loss function. An improved version of the above loss function selects an average of multiple $k_i$ values that are within $\delta$ (we use $\delta$ from $1\%$ to $3\%$) of the $T$th percentile:
    \begin{equation}
        k_{B} = \frac{\sum_{i \in B} c_i k_i}{\sum_{i \in B}{c_i}}
    \quad
    \text{where}
    \quad
    c_i =
    \begin{cases}
    1 & \text{if $k_i$ is within $\delta$ of the $T$th percentile of $B$} \\
    0 & \text{otherwise} \\
    \end{cases}
    \end{equation}

\subsection{Training Procedure and Testing} \label{pretraining}

We have found that to get the best results training EIM models, one should use larger minibatches than in other techniques. Since EIM uses the minibatch to estimate its coverage, larger minibatches produce more stable estimates that improve convergence and performance. Another procedure we found profitable was to pretrain EIM on fixed bounds. This pretraining uses mean squared error loss on the target value plus and minus a constant on a minibatch $B$, with the following pretraining loss function:
\begin{equation}
E_{B}' = \sum_{i \in B} (y_i - \alpha - l(x_i))^2 + \sum_{i \in B} (y_i + \alpha - u(x_i))^2
\end{equation}
Where $y_i$ is target value, $[l(x_i),  u(x_i)]$ is the predicted range, and $\alpha$ is a constant depends on the dataset. This pretraining was used mainly for preventing early divergence when training with the EIM loss. To use EIM in a production environment the output bounds from the trained model should be scaled about their center in a similar process to how $k_{b}$ is computed. After training the model, we use a holdout set to compute the population scaling factor $k$ required to hit the PICP target. We then grow or shrink all future model outputs by this constant factor $k$ (see section \ref{scaledintervals}).

\section{Evaluation Datasets}

\begin{figure}[ht]
\centering
\resizebox{.6\linewidth}{!}{
    \begin{tabular}{ r | c | c} 
     Property                  & Domain Valuation Dataset & Million Song Dataset \\
     \hline
     Size                      & 634,328      & 515,345    \\ 
     Target range              & \$0-\$25,000 & 1922-2011  \\ 
     Target mean               & \$1,971.2    & 1998.4   \\ 
     Target median             & \$1,000.0    & 2002  \\
     Target standard deviation & \$3,379.8    & 10.93 \\ 
     Number of attributes      & 269          & 90 \\
     Base model $R^2$          & 56.83        & 20.67
    \end{tabular}}
    \caption{Properties of each dataset. \label{fig:dataset}}
\end{figure}

This section describes the two real world datasets used to evaluate EIM.  Figure~\ref{fig:dataset} shows some summary statistics about each of these datasets.

\subsection{Aftermarket Domain Valuation Dataset} \label{valuation_dataset}

For our first dataset, we use a set of aftermarket sale prices of domain names. The goal of our models are: given a domain name that has sold in the past, provide prediction intervals on the price at which it sold.  This is a real-world dataset that is being used to build models available to millions of customers. A challenge in working with domain names is tokenizing them into words, because domains do not contain the spaces found in most text.  We built a domain name tokenizer based on word embeddings~\cite{mnih2013learning} and language model that estimates the probability of every possible tokenization. We built a training set for this model by extracting the tokenization of a domain name from its crawled website content. This language model for domains is also used as an input to our neural network, as we find the way people use words in domain names differs from how they use them in other text.

Our model also pulls in external data and other features as inputs to the neural network. These additional inputs include:
\begin{itemize}

\item For each top-level domain (TLD) with enough historical sale data, we create a vector embedding. The TLD is one of the most important features. 
\item The usage of other TLDs of the domain (e.g., when looking at $foo.com$, we inspect $foo.net$ and others to see which hosting provider, if any, serves it.).
\item Domain statistics from other aftermarket datasets, such as active listings and expiry auctions.
\item Word matches and statistics from various dictionaries, including English, English part-of-speech, French, German, Italian, Spanish, Japanese, first names, last names, female names, male names, places, acronyms, brands, products, adult-related, currencies, phrases, countries, and Wikipedia.
\item Other handcrafted features to identify specific non-word based patterns of interest to domain investors~\cite{chips}.
\end{itemize}

\subsection{Year Prediction Million Song Dataset} \label{msd}

The second dataset we used is a published dataset~\cite{Dua:2017} that is a subset of the Million Song Dataset~\cite{Bertin-Mahieux2011}. This dataset is used for building regression models that predict the release year of a song from audio features. Each instance has 90 attributes which consist of 12 timbre averages and 78 timbre covariances encoded as floats.

\subsection{Scaled Intervals} \label{scaledintervals}
Some of the models we compare against are not able to customize themselves to a specific target correctness rate, or may be under or over the target PICP.  To provide a fair playing field and avoid comparing models with different PICP rates, we scale the output of each model so it exactly hits the target ($PICP=T$). To do this, we define a new interval $[u'(x_i), l'(x_i)]$ based on the raw interval $[u(x_i), l(x_i)]$: 

\begin{equation}
\begin{aligned}
u'(x_i) &= \frac{u(x_i) + l(x_i) + k |u(x_i) - l(x_i)|}{2} \\
l'(x_i) &= \frac{u(x_i) + l(x_i) - k |u(x_i) - l(x_i)|}{2}
\end{aligned}
\end{equation}

where $k$ is a constant scaling factor.  Note that if $k=1$, then $[u'(x_i), l'(x_i)] = [u(x_i), l(x_i)]$.  If $k<1$, the intervals are shrunk and if $k>1$, they are expanded in a linear way.  We compute the $k$ required to bring each model into exact PICP compliance.  

\subsection{Dataset Specific Neural Network Structures} \label{experiment_setup}

\begin{figure}[!htb]
    \centering
    \subfigure[Single Output +\newline Domain Valuation\label{subfig:sval}]{
        \includegraphics[width=.25\linewidth]{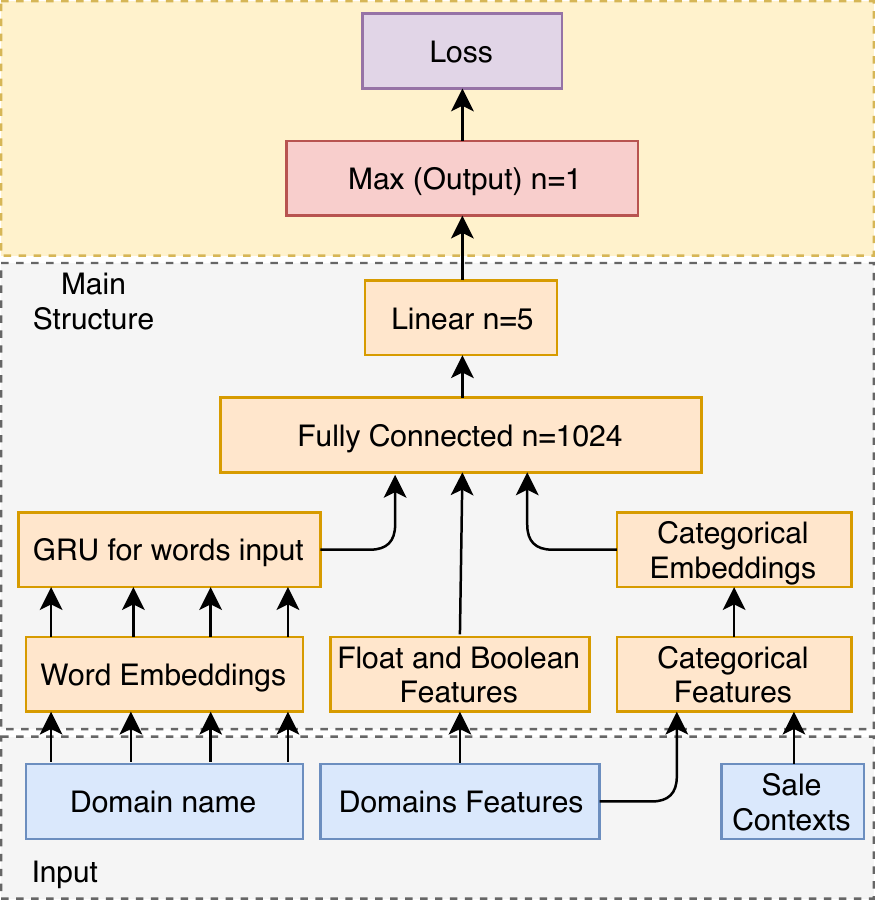}}\hfill
    \subfigure[Dual Output +\newline Domain Valuation\label{subfig:dval}]{
        \includegraphics[width=.25\linewidth]{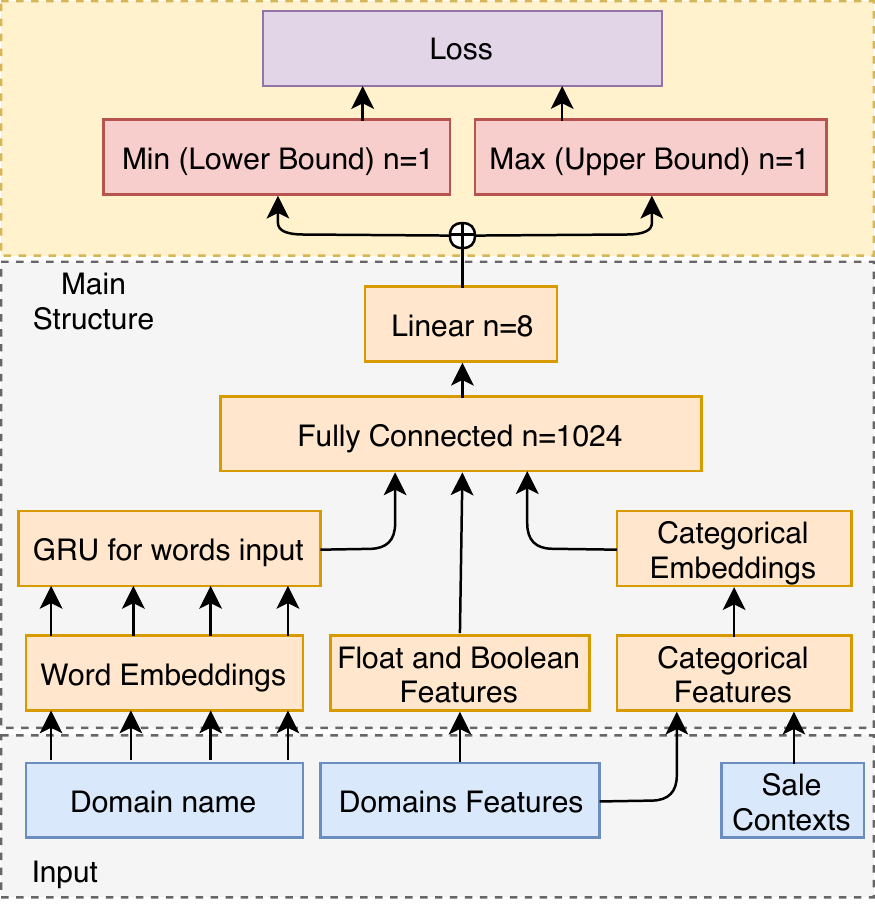}}\hfill
    \subfigure[Single Output +\newline Million Song\label{subfig:smsd}]{
        \includegraphics[width=.237\linewidth]{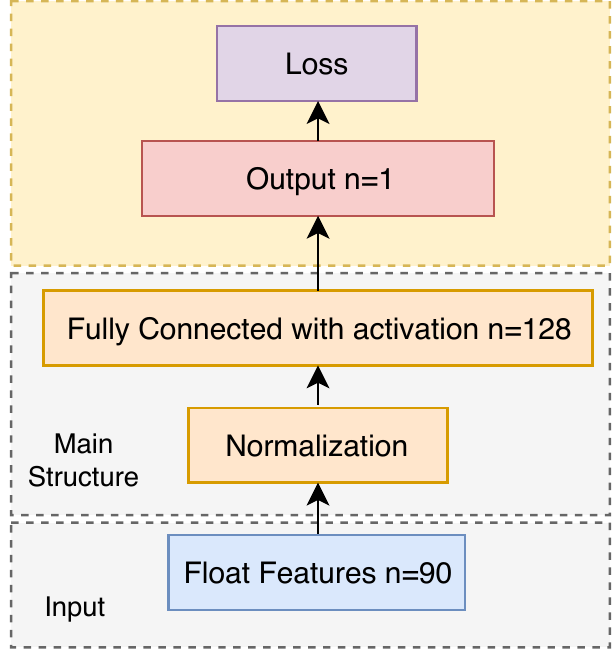}}\hfill
    \subfigure[Dual Output +\newline Million Song\label{subfig:dmsd}]{
        \includegraphics[width=.237\linewidth]{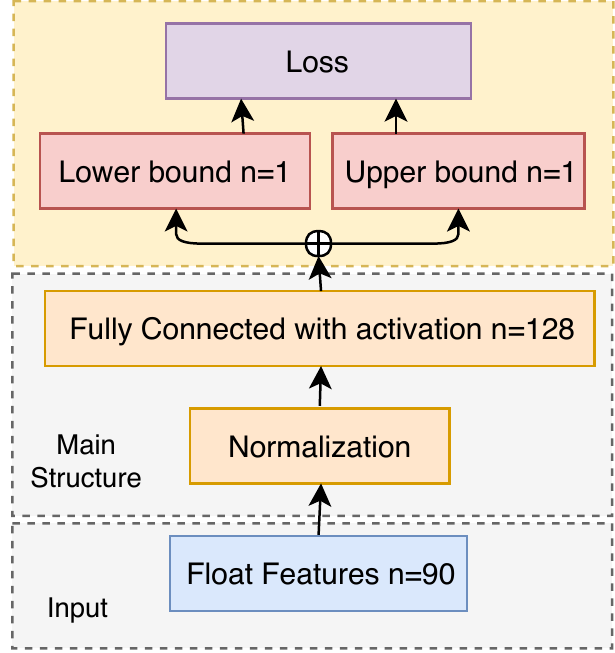}}
    \caption{
    Neural network structures for each dataset. 
    Figures~\ref{subfig:sval} and~\ref{subfig:smsd} show the structures used by the techniques fixed bounds, MLE, and ensemble.
    Figures~\ref{subfig:dval} and~\ref{subfig:dmsd} show the structures used by EIM and quantile regression.
    \label{fig:structure}}
\end{figure}

To compare the performance of different methods on the Domain Valuation Dataset and the Million Song Dataset, we first implemented the regression models that can output a single target value given the input features. Figures~\ref{subfig:sval} and figures~\ref{subfig:smsd} show the single output neural network structures for both the Domain Valuation Dataset and Million Song Dataset, respectively. Note that Figure~\ref{subfig:sval} is a more complex model because of the features introduced in section \ref{valuation_dataset}. Figure~\ref{subfig:dval} and Figure~\ref{subfig:dmsd} show the neural network structures based on the single neural network structure but modified to output both lower and upper bound predictions.

For our implementations, maximum likelihood method used two networks to predict the target value and the total prediction variance, respectively. Ensemble method used 200 networks to obtain the mean prediction and the model uncertainty variance, and trained an additional network to predict the data noise variance. Both quantile regression method and EIM method used only one network to output both lower and upper bounds, and we built 3 different models for 3 different PICP targets $70\%$, $80\%$ and $90\%$. For EIM, we can specify different PICP targets in the loss function for different models; for quantile regression method, we run an exhaustive grid search to find the best possible quantiles range ($\tau_l$, $\tau_u$) for different PICP targets.

\section{Results and Discussion} \label{results}

\begin{figure}[t]
    \subfigure[MPIW for Domain Valuation Dataset\label{subfig:val}]{\resizebox{.47\linewidth}{!}{
        \begin{tabular}{r | c | c | c }
         Method & PICP=70\% & PICP=80\% & PICP=90\% \\
         \hline
         MLE                 & 2488 & 2735 & 3689 \\ 
         Ensemble            & 1908 & 2480 & 3437 \\ 
         EIM 70              & \textbf{1185} & 1889 & 3715 \\
         EIM 80              & 1452 & \textbf{1692} & 3076 \\
         EIM 90              & 2183 & 2362 & \textbf{2756} \\
         Quantile 70         & 1804 & 2509 & 4802 \\
         Quantile 80         & 1969 & 2402 & 3519 \\
         Quantile 90         & 2108 & 2562 & 3485 \\
         Fixed bounds        & 2642 & 3077 & 3654 \\
        \end{tabular}}}
    \hfill
    \subfigure[MPIW for Million Song Dataset\label{subfig:msd}]{\resizebox{.47\linewidth}{!}{
        \begin{tabular}{r | c | c | c }
         Method & PICP=70\% & PICP=80\% & PICP=90\% \\
         \hline
         MLE                 & 15.50 & 18.93 & 27.37 \\ 
         Ensemble            & 16.92 & 19.91 & 25.74 \\ 
         EIM 70              & \textbf{11.54} & 16.53 & 32.30 \\
         EIM 80              & 12.18 & \textbf{15.00} & 26.23 \\
         EIM 90              & 14.81 & 17.35 & \textbf{21.32} \\
         Quantile 70         & 12.59 & 16.40 & 27.39 \\
         Quantile 80         & 12.70 & 15.99 & 25.47 \\
         Quantile 90         & 14.80 & 17.50 & 22.46 \\
         Fixed Bounds        & 23.57 & 26.20 & 28.96 \\
        \end{tabular}}}
        
    \caption{Mean prediction interval width (MPIW) at 70\%, 80\%, and 90\% prediction interval coverage percent (PICP) for all techniques and both datasets.  EIM and quantile regression are parameterized by each target PICP, where EIM 80 indicates EIM trained to hit a PICP=80\%.  Lower is better and bold values are the best found.
    \label{fig:mainresults}}
\end{figure}

Figure~\ref{fig:mainresults} shows that EIM produces significantly tighter prediction intervals than other techniques for both the Domain Valuation Dataset and the Million Song Dataset.
EIM produces $1.33x$ to $2.23x$ tighter bounds than fixed bounds method,
$1.26x$ to $2.1x$ tighter bounds than MLE, 
$1.21x$ to $1.61x$ tighter bounds than ensemble, 
and $1.06x$ to $1.52x$ tighter bounds than quantile regression.
Interestingly the gap between EIM and the next best technique decreases with higher PICP targets,
going from $1.26x$ tighter prediction intervals for a 70\% coverage target to $1.06x$ a 90\% target. We believe that these lower targets provide EIM  more flexibility in the choice of generated ranges which it is better able to utilize than other techniques.

\begin{figure}[ht]
 \centering
 \includegraphics[width=.4\linewidth]{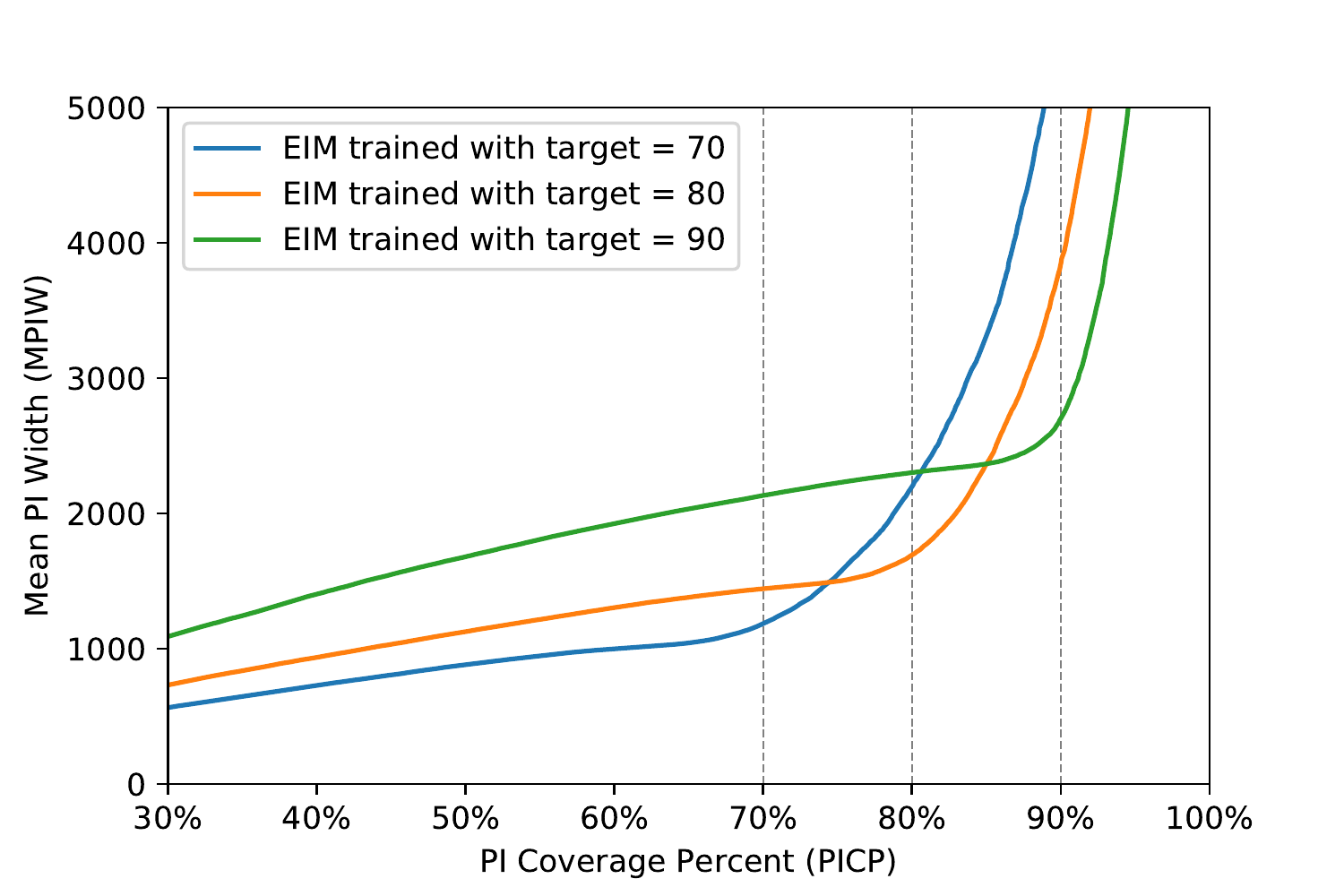}
 \caption{MPIW for the versions of EIM trained for one PICP target scaled to hit other PICP targets on the domain valuation dataset.\label{fig:EIMdist}}
\end{figure}

Figure~\ref{fig:mainresults} also shows that techniques that train for the specific PICP targets (EIM and quantile regression) have a significant advantage over the techniques that only support symmetric bounds (fixed, MLE, and ensemble).  One can note that the 
versions of these techniques with training targets matching the measured target perform better than those trained with other targets.
Figure~\ref{fig:EIMdist} expands on this finding for EIM by showing how MPIW changes as one scales versions EIM trained for one PICP target to other PICP targets on the Domain Valuation Dataset.  We can see that each version of EIM has specialized itself for its specific target.

\begin{figure}[ht]
    \def\figheight{8em}
    \subfigure[Domain Valuation, PICP=70\%]{\includegraphics[width=.32\linewidth,height=\figheight]{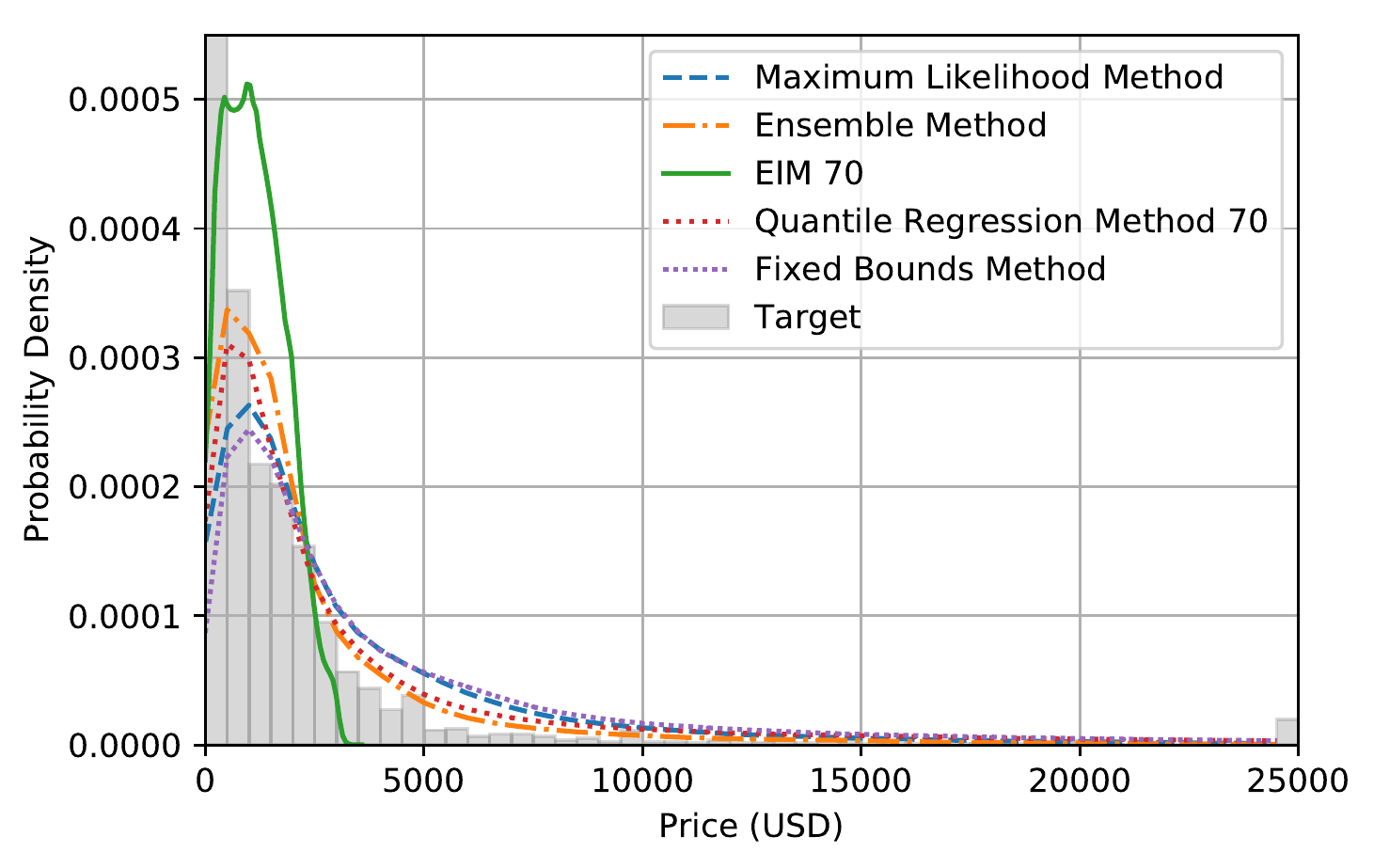}}
    \subfigure[Domain Valuation, PICP=80\%]{\includegraphics[width=.32\linewidth,height=\figheight]{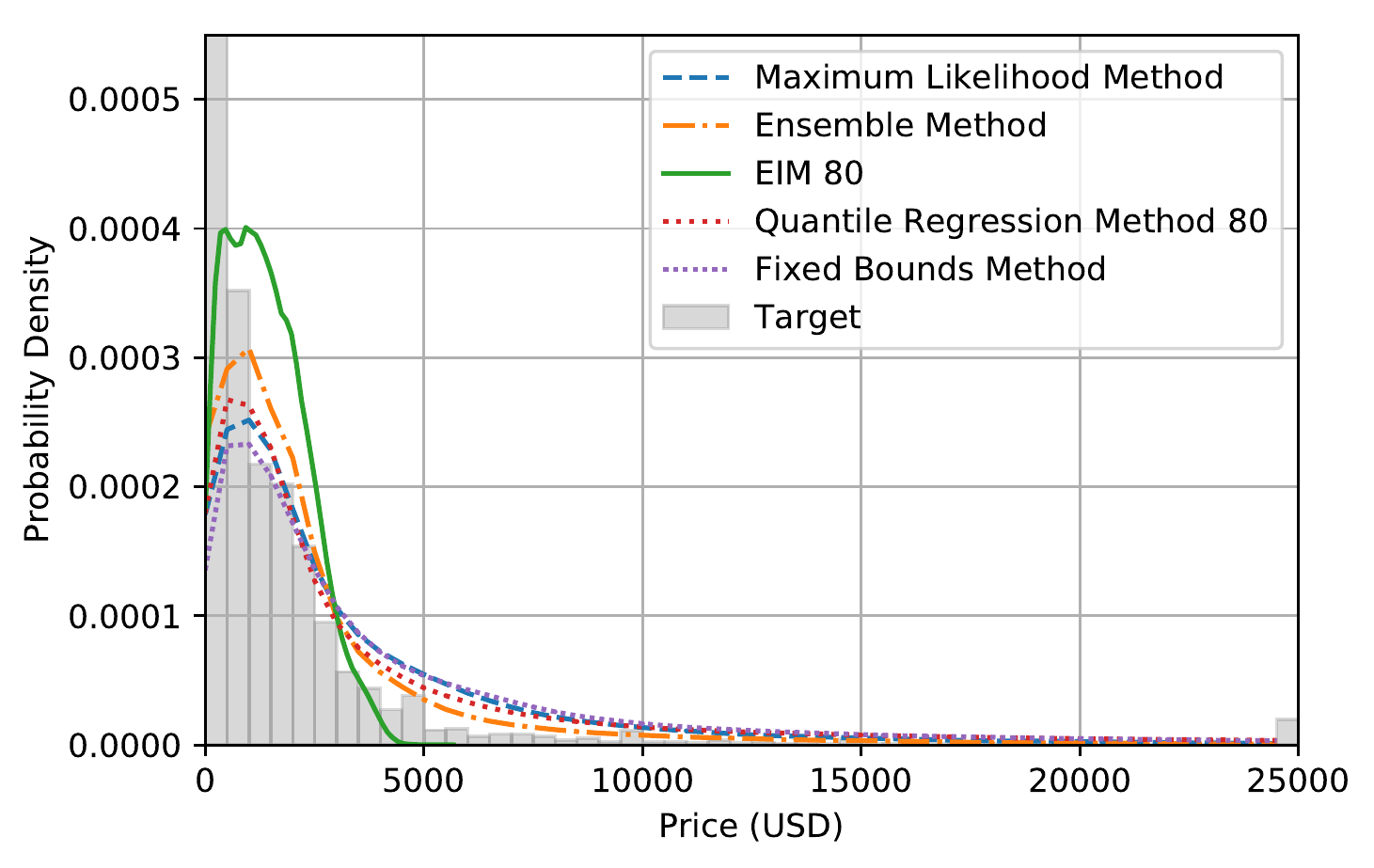}}
    \subfigure[Domain Valuation, PICP=90\%]{\includegraphics[width=.32\linewidth,height=\figheight]{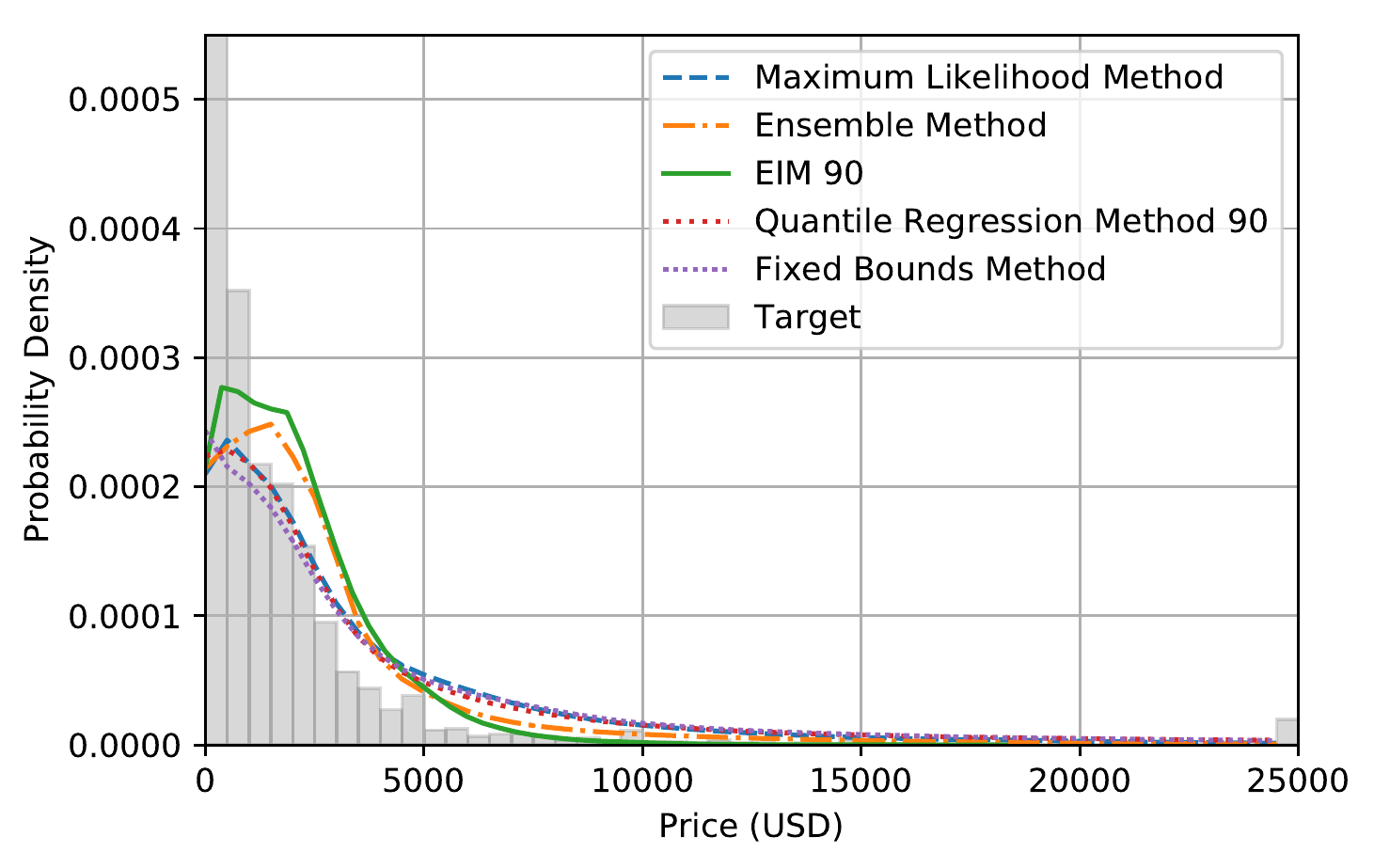}}
    \subfigure[Million Song, PICP=70\%]{\includegraphics[width=.32\linewidth,height=\figheight]{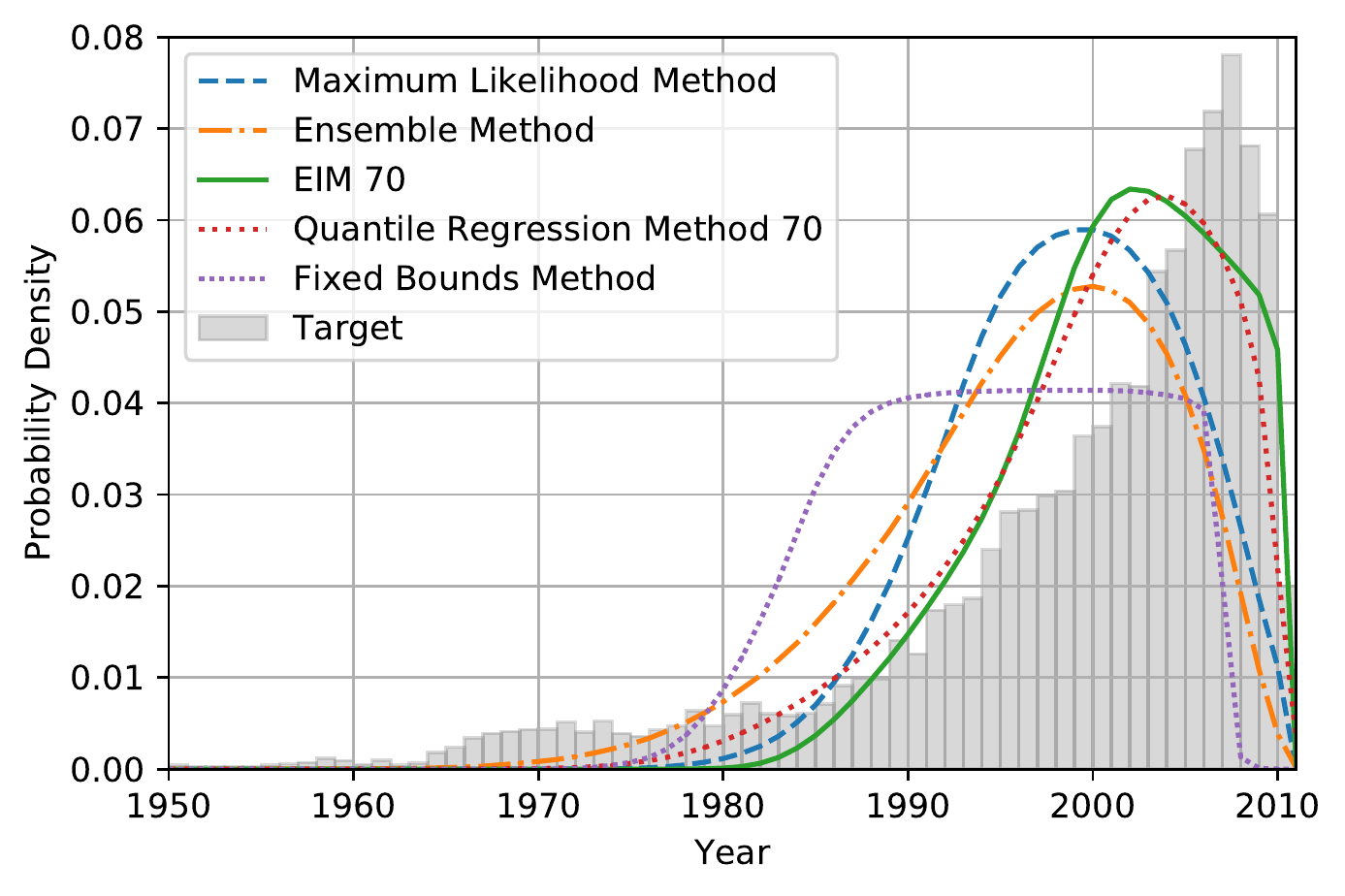}}
    \subfigure[Million Song, PICP=80\%]{\includegraphics[width=.32\linewidth,height=\figheight]{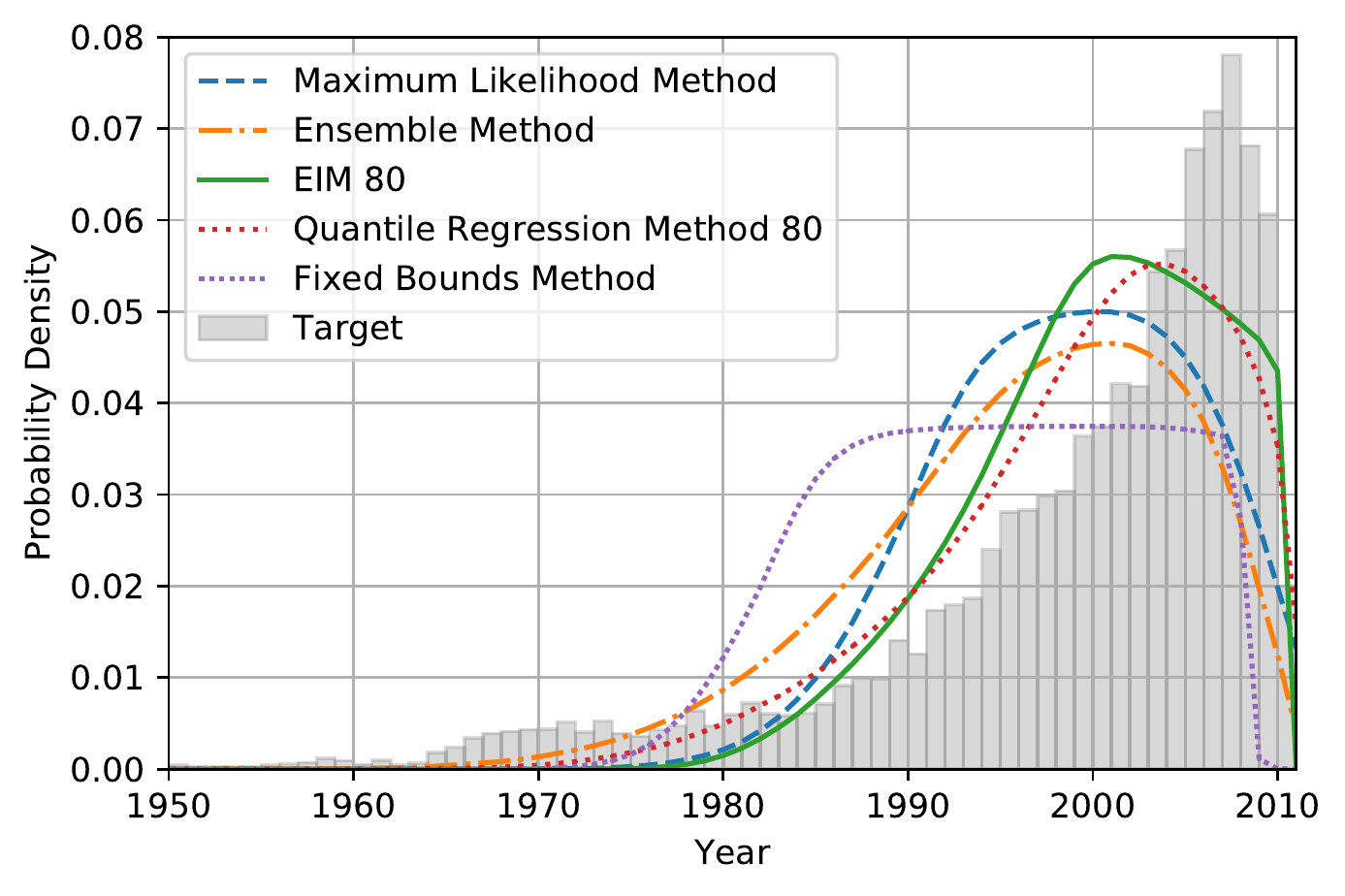}}
    \subfigure[Million Song, PICP=90\%]{\includegraphics[width=.32\linewidth,height=\figheight]{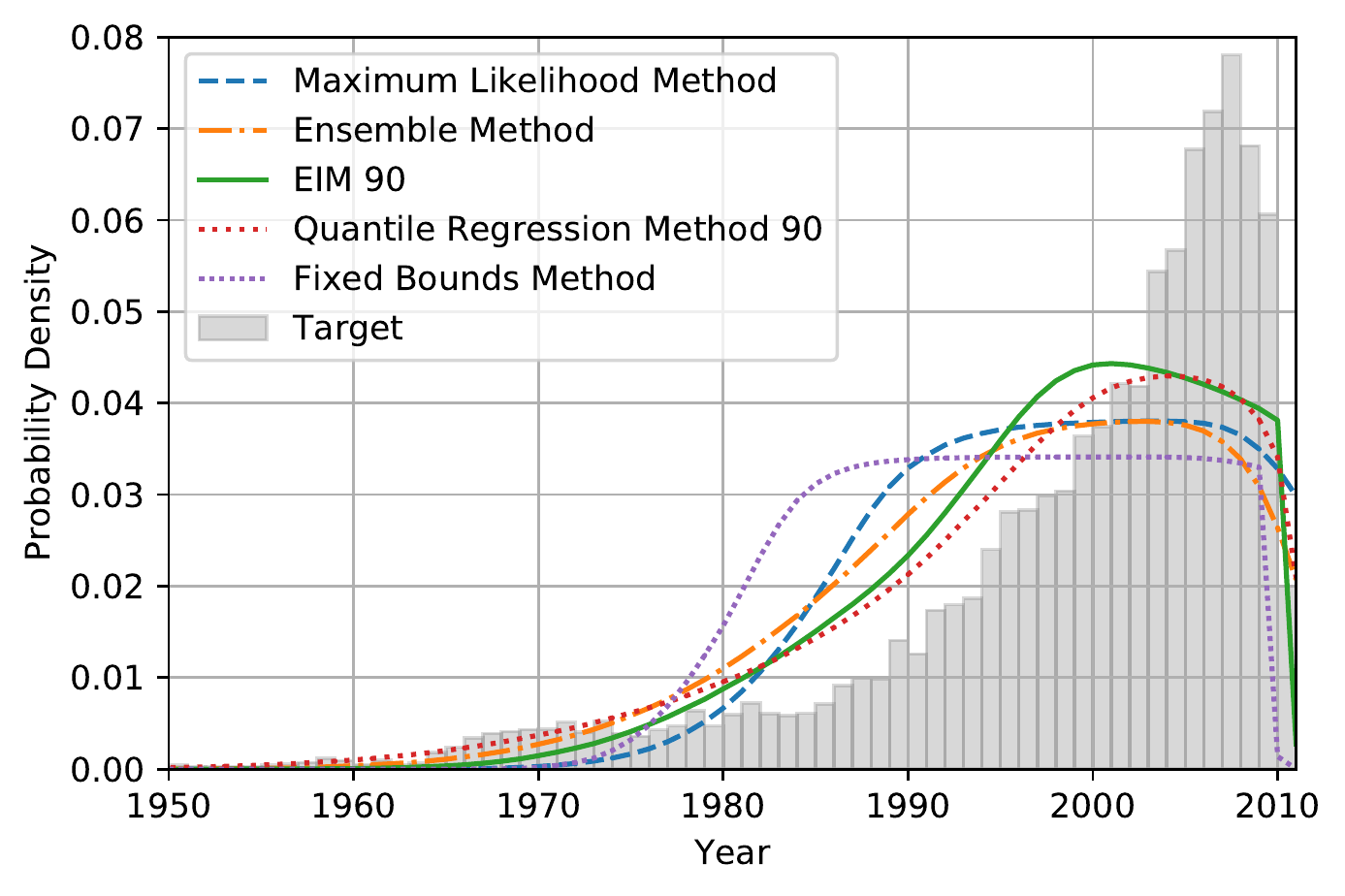}}
    \caption{Prediction and target distribution for each target and dataset.\label{fig:distribution} Note that the first bar in (a)-(c) is 0.00068, and was truncated for clarity in showing the predicted distributions.}
    \vspace{-1em}
\end{figure}

Figure~\ref{fig:distribution} shows the prediction and target distribution for each target and dataset.  The target distribution is a histogram of actual values in the holdout set.  The lines for each technique show the normalized probability density corresponding to how often the predicted range contains each point.   

Looking at Figure~\ref{fig:distribution}(d-f) one notices that the predicted distributions for EIM and quantile regression are skewed more towards where the majority of the data is compared to the other techniques.  The other methods construct their bounds by starting at the predicted mean and expanding equally in both directions.  EIM, on the other hand, can strategically expand bounds more in the direction with more density of data.  For Figure~\ref{fig:distribution}(a-c), the data distribution is more concentrated at the lower values, and the prediction distribution of EIM stands out from all the other techniques in that it is able to change its distribution more dramatically between the different target percentiles.

Comparing the three target percentages in Figure~\ref{fig:distribution} one sees that the techniques need to spread out their prediction distributions to achieve higher coverage percentages.  One also sees a shift in the mean which is more pronounced for EIM.  We see larger differences between techniques for the lower PICPs target, and all the techniques cluster closer together at the higher targets.

Finally, we should mention the different complexities of the top three techniques: EIM, quantile, and ensemble.  EIM is by far the fastest and most straightforward of them, requiring training only a single neural network.   Ensemble is the slowest and most complex, requiring training 200 neural networks and using sophisticated but challenging-to-implement methods to combine them.  Basic quantile regression has a certain elegance to it but contains two hyperparameters which must be set with exhaustive grid search to achieve optimal results.  This grid search made the implementation more complicated and made quantile regression the slowest technique to train. The resulting values for lower and upper quantiles found by the grid search were non-obvious.  For example, for the Million Song Dataset they were (0.2, 0.65) for PICP=70\%, (0.2, 0.75)  for PICP=80\%, and (0.5, 0.75) for PICP=90\%.  When scaled to hit the coverage, these outperformed the more naive symmetric choices.

\section{Conclusions}

This paper presented Expanded Interval Minimization (EIM), a novel technique for generating prediction intervals with neural networks.   We showed that compared to the next best technique, EIM produces $1.26x$ tighter prediction intervals for a 70\% coverage target, $1.21x$ tighter prediction intervals for an 80\% target, and $1.06x$ tighter intervals for a 90\% coverage target. EIM is a natural fit for any application using prediction intervals and having asymmetrically distributed error.  Using EIM, we hope that others will be able to generate tighter prediction intervals and advance the state of the art of what machines can do with deep learning. 

\bibliography{nnpi}
\bibliographystyle{abbrv}

\end{document}